# Study of Dropout in PointPillars for 3D Object Detection

Xiaoxiang Sun

xiaoxianglucky2023@163.com

Geoffrey Fox

gcfexchange@gmail.com

## Abstract

3D object detection is critical for autonomous driving, leveraging deep learning techniques to interpret LiDAR data. The PointPillars architecture is a prominent model in this field, distinguished by its efficient use of LiDAR data. This study provides an analysis of enhancing the performance of PointPillars model under various dropout rates to address overfitting and improve model generalization. Dropout, a regularization technique, involves randomly omitting neurons during training, compelling the network to learn robust and diverse features. We systematically compare the effects of different enhancement techniques on the model's regression performance during training and its accuracy, measured by Average Precision (AP) and Average Orientation Similarity (AOS). Our findings offer insights into the optimal enhancements, contributing to improved 3D object detection in autonomous driving applications.

## 1. Introduction

3D object detection is a significant component of autonomous driving, heavily relying on deep learning techniques. Among various models, Lang's PointPillars[1] architecture stands out as a distinguished 3D object detection model, based on Pytorch[22]. It prominently utilizes LiDAR data, setting it apart from traditional camera-based methods.

Dropout plays a crucial role in preventing overfitting in neural networks. This article focuses on evaluating the performance of the enhanced PointPillars model under various dropout rates during training.

Dropout involves randomly "dropping out" a subset of neurons during training. During each forward and backward pass, each neuron has a probability of being temporarily removed from the network, forcing the network to learn redundant representations and not rely too heavily on any single neuron. The main significance of dropout lies in its ability to improve the generalization of the model. By randomly omitting neurons during training, dropout prevents the network from becoming too reliant on specific neurons, which can lead to overfitting. This enhancement helps the model perform better on unseen data, as it learns more robust and diverse features.

The performance of dropout depends on the probability rate of temporarily removing a neuron from the network, known as the dropout rate. This article explores different dropout rates and evaluates their respective performances. Two aspects of evaluation are proposed to analyze the model in detail. The first is the regression performance of the training process, with loss utilized to reflect this property. The second is the accuracy of the trained model[2], reflected by metrics such as Average Precision (AP)[10] and Average Orientation Similarity (AOS)[7].

In this article, the methodology of the algorithms is introduced, including the methods used to set dropout during model training and the evaluation algorithms. Subsequently, the results based on different dropout rates are presented and analyzed. The main contributions of this article can be summarized as follows.

1. Introducing a comprehensive evaluation of PointPillars with various dropout rates.
2. Analyzing the impact of dropout on both the training process and the accuracy of the trained model.
3. Providing insights into the optimal dropout rates for enhancing the performance of 3D object detection models using LiDAR data.

## 2. Related Work

3D object detection[3-6,23-27] is a fundamental task in the perception system of autonomous vehicles[21], allowing them to understand their environment and make safe navigation decisions. LiDAR[3] provides detailed spatial information by emitting laser beams and measuring the time it takes for them to reflect off surfaces.

Voxel-Based Methods, such as VoxelNet[4], divides the point cloud data into a grid of voxels (3D pixels) and process each voxel to extract features. This structured representation allows for efficient processing using 3D convolutional neural networks (CNNs). Point-Based Methods, such as PointNet[5] and its variants[3] process raw point clouds directly without converting them into a voxel grid. These methods use neural networks to learn features from the raw point data, which can capture fine-grained details and spatial relationships. Hybrid approaches[6] combine the strengths of both voxel-based and point-based methods. They use voxelization to provide a structured representation while retaining the ability to process raw point data for finer details. Methods like Point-Voxel CNNs (PVCNN) leverage this hybrid approach to improve detection accuracy and efficiency.

Recent advancements in 3D object detection have focused on improving the robustness and accuracy of detection models in various driving scenarios. For example, techniques like data augmentation, domain adaptation, and adversarial training have been employed to enhance the generalization capabilities of these models[3]. Moreover, the use of large-scale datasets, such as KITTI[7], nuScenes[8], and Waymo Open Dataset[9], has facilitated the development and benchmarking of more sophisticated models. This article focuses on enhancing the accuracy of detection models by changing different dropout, based on KITTI dataset.

Performance evaluation of 3D object detection models is crucial for their deployment in autonomous driving. Key metrics include Average Precision (AP) and Intersection over Union (IoU)[10], which measure the accuracy and overlap between detected objects and ground truth annotations. Robustness to varying conditions, such as different weather and lighting, is also an important aspect of evaluation

The integration of 3D object detection into autonomous driving systems enhances the vehicle's ability to perceive and interact with its environment safely and effectively. Continuous advancements in sensor technology, data processing algorithms, and machine learning models are driving the progress in this field, aiming to achieve higher levels of safety and autonomy in self-driving vehicles.

Dropout is a widely used regularization technique designed to improve the generalization capabilities of neural networks by preventing overfitting during the training phase. Srivastava[11] introduced dropout as a method to prevent co-adaptation of feature detectors in neural networks, demonstrating its effectiveness across various tasks, including image classification and speech recognition. Their work laid the foundation for subsequent studies on dropout and its variants. Salehin and Kang[12] provided a comprehensive review of dropout regularization approaches, discussing various methods such as standard dropout, DropConnect[13], and DropBlock[14]. They highlighted how these techniques adapt to different neural network architectures, such as CNNs and RNNs, to enhance model performance. Their review underscored the importance of tailoring dropout techniques to the specific needs of the model architecture. Khan[15] proposed spectral dropout, which selectively drops frequencies in the neural network layers. This method was particularly effective in regularizing convolutional neural networks, demonstrating improved robustness and generalization compared to standard dropout techniques. Hinton[16] initially proposed dropout and highlighted its role in improving the generalization of deep networks by preventing the network from becoming too reliant on specific neurons. Their work showed significant performance improvements in various machine learning tasks, setting a precedent for further research on dropout and its impact on neural network training.

In the context of 3D object detection for autonomous driving, dropout techniques have been evaluated to understand their impact on model performance. Studies have shown that proper tuning of dropout rates can significantly enhance the model's ability to detect objects accurately in diverse and dynamic driving environments[17]. This highlights the critical role of dropout in improving the generalization of models used in safety-critical applications.

The PointPillars model, introduced by Lang et al., has become a prominent framework for 3D object detection using LiDAR data. It converts point cloud data into a structured pseudo-image representation, enabling the use of 2D convolutional neural networks for feature extraction and object detection. This architecture has been praised for its efficiency and high performance in real-time applications. Based on the architecture, researches have been done to enhance its performance[18-20].

ET-PointPillars[18] focus on enhancing the PointPillars architecture through optimized voxel downsampling. This enhancement is shown to improve the efficiency and accuracy of the model by reducing computational overhead without compromising performance. Although the study does not specifically target dropout, it provides valuable insights into overall model enhancement strategies that contribute to robust 3D object detection. Implementation of the PointPillars Network in Reprogrammable Heterogeneous Devices[19] explores the optimization of the PointPillars network for embedded systems, highlighting performance improvements and trade-offs in accuracy. The study examines how the integration of hardware and software can enhance the model's efficiency.

# 3. Methods

## 3.1 Dropout Methods

The PointPillars architecture for 3D object detection involves several key stages, as shown in Fig.1. First, the point cloud is divided into vertical columns called pillars, with each pillar containing a set of points (voxelization). In the Pillar Feature Encoder stage, point features are projected to a higher-dimensional space using a linear layer. The Convolutional Backbone then applies a series of convolutions, along with batch normalization and ReLU activations, to extract spatial features from the pseudo-image formed by the pillars. This is followed by a max-pooling layer to create a fixed-size representation for each pillar. Then, a scatter operation aggregates these features within each pillar.

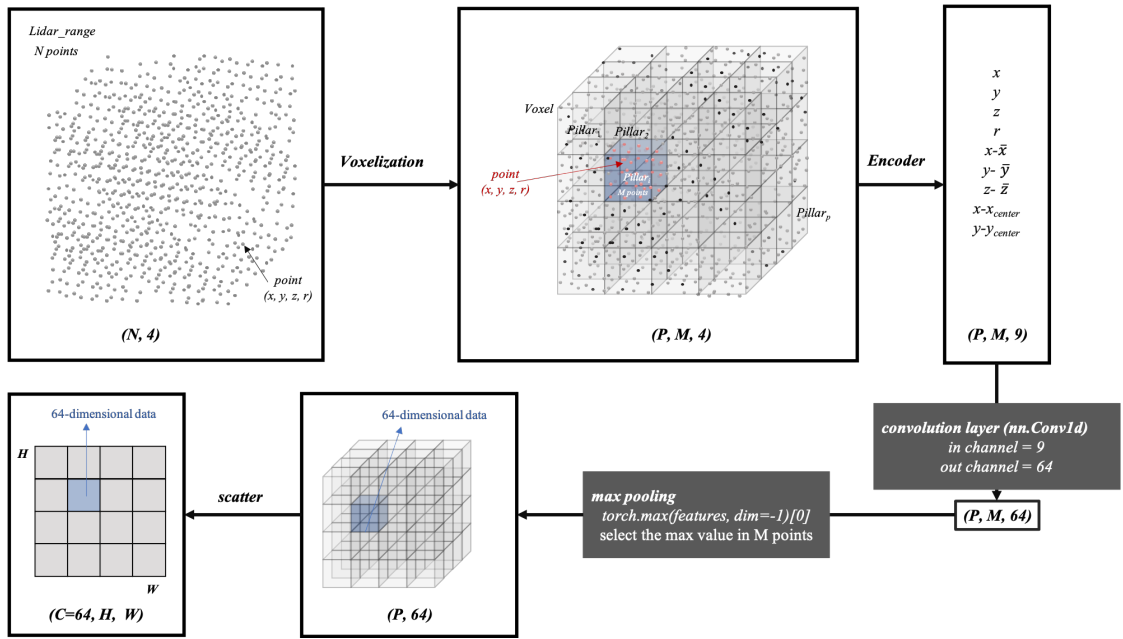

Fig.1　　Layers of PointsPillars.

In this article, we enhance the PointPillars architecture by incorporating dropout into the convolutional layers. Specifically, dropout is added after each convolutional layer in the network. This integration involves inserting dropout layers, which randomly deactivate a subset of neurons with a specified probability during each training iteration. By placing these dropout layers immediately after the convolutional layers, we ensure that the feature maps generated by the convolutions are regularized, which helps in reducing overfitting.

## 3.2 Loss Function

Loss function is utilized to evaluate a aspect of the performance of trained model in this article, including classification loss, regression loss and orientation classification loss.

1. Classification Loss: Focal Loss
To mitigate the class imbalance between positive and negative samples in our dataset, we utilize Focal Loss[3] as classification loss. The formulation of Focal Loss is

$$FL(p_t) = -\alpha_t(1 - p_t)^\gamma \log(p_t)$$

$$\alpha_t = \begin{cases} \alpha, y = 1 \\ 1 - \alpha, y = 0 \end{cases}, p_t = \begin{cases} p, y = 1 \\ 1 - p, y = 0 \end{cases}$$

where, $\alpha_t$ is a scaling factor that balances the importance between positive and negative samples; $p_t$ represents the predicted probability of the true class.

In the implementation of PointPillars, we set $\alpha = 0.25$ and $\gamma = 2.0$. These hyperparameters are crucial as they reduce the loss contribution from easy examples, thereby emphasizing the learning on hard examples. The impact of Focal Loss on classification loss is shown in the table below, which demonstrates a significant reduction in the influence of the numerous negative anchors, thus enhancing the model's focus on the minority positive anchors.

Table.1 The impact of Focal Loss on classification loss.

| Positive Anchors | Negative Anchors | Positive Anchors Loss (without FL) | Negative Anchors Loss (without FL) | Positive Anchors Loss (with FL) | Negative Anchors Loss (with FL) |
|---|---|---|---|---|---|
| 938 | 5780509 | 4416.1719 | 58345.1328 | 1083.7280 | 4.7090 |

2. Regression Loss: SmoothL1 Loss

For bounding box regression, we employ Smooth L1 Loss[4], which is less sensitive to outliers compared to the standard L2 loss[5]. This loss function combines the advantages of L1 and L2 losses, depending on the magnitude of the error. The mathematical expression is

$$L(x_n, y_n) = \begin{cases} 0.5(x_n - y_n)^2/\beta, |x_n - y_n| < \beta \\ |x_n - y_n| - 0.5\beta, otherwise \end{cases}$$

In this study, we set $\beta = 1/9$. Smooth L1 Loss penalizes large deviations linearly and small deviations quadratically, thereby offering a balanced and robust error measurement.

3. Orientation Classification Loss: Cross Entropy Loss

For orientation classification, we use the Cross Entropy Loss, a standard loss function for classification tasks that outputs a probability value between 0 and 1. The formula is

$$L(y, \hat{y}) = -\sum_{i=1}^{n} \hat{y}_i \log(y_i)$$

where, $\hat{y}$ denotes the predicted probability distribution, while $y$ represents the ground truth distribution. This loss function is particularly effective for handling multi-class classification problems.

4. Total Loss Calculation

The overall loss for training the model is computed as a weighted sum of the aforementioned loss components:

$$total\_loss = 1.0 \times classification\_loss + 2.0 \times regression\_loss + 2.0 \times orientation\_classification\_loss$$

**3.3  Accuracy Evaluation Method**

AP and AOS are utilized to evaluate the accuracy performance of trained model in this article. Precision-Recall (PR) curve can be used for qualitative analysis of model accuracy in detection problem of multiple categories of objects, and Average Precision (AP) can be used for quantitative analysis of model accuracy. For object orientation detection, Average Orientation Similarity (AOS) can be used to measure the similarity between the detection results and the ground truth orientation. The definition, usage and calculated methods are introduced as follows.

In this paper, another feature is proposed to evaluate different types of AOS and AP, named *difficulty*. Based on the height of the 2D box, the level of occlusion and truncation, the bounding boxes (bbox) are divided into $difficulty = 0, 1, 2, -1$. The definition rules are set as follows, where height, occlusion and truncation are labeled by KITTI[7].

1. if $height > 45$, $occlusion \leq 0$, and $truncation \leq 0.15$, set $difficulty = 0$:
2. if $25 < height \leq 45$, $0 < occlusion \leq 1$, and $0.15 < truncation \leq 0.3$, set $difficulty = 1$;
3. if $25 < height \leq 45$, $1 < occlusion \leq 2$, and $0.3 < truncation \leq 0.5$, set $difficulty = 2$;
3. if other conditions, set $difficulty = -1$.

Precision is the percentage of *True Positive* samples among all positive samples given by the detector, to evaluate the accuracy of the detector in detecting success. It is calculated by

$$Precision = \frac{TP}{TP + FP}$$

where, $TP$ represents the number of True Positive samples calculated by the detector correctly, and $FP$ represents the number of False Positive samples calculated by the detector incorrectly.

Recall is the percentage of *True Positive* samples among all positive samples given by the ground truth, to evaluate the detector's coverage rate of all targets to be detected. It is calculated by

$$Recall = \frac{TP}{TP + FN}$$

where, $FN$ represents the number of False Negative samples calculated by the detector incorrectly.

PR curve is a curve with Precision as the vertical axis and Recall as the horizontal axis. By selecting different confidence thresholds, different points can be obtained on the PR coordinate system, and connecting these points yields the PR curve, as shown in Fig.1. The curve is to evaluate model performance. The larger the Precision and Recall values, the better, so the PR curve should ideally protrude towards the top right corner.

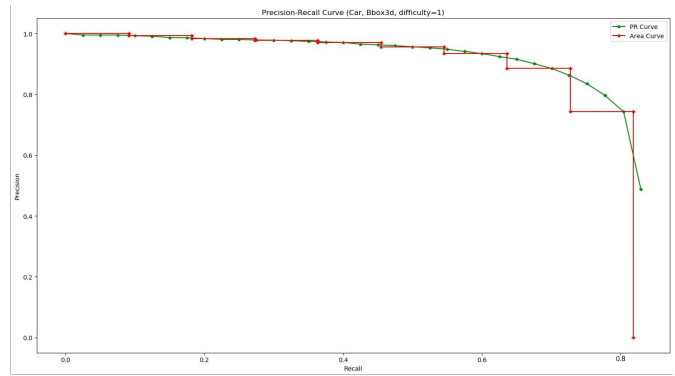

Fig.2　　The example of PR curve.

AP is the area under the PR curve. It can be estimated using the 11-point interpolation method and the all-points interpolation method. AP is to measure the average precision of the algorithm for a single category. The higher the AP value, the higher the detection accuracy for that category. There are 3 calculation methods of AP corresponding to 3 different IoU calculation methods, named AP(2D), AP(3D), AP(BEV) respectively. IoU is the ratio of the intersection to the union of the predicted box and the ground truth box, to judge the overlap degree of two boxes. The higher the value, the higher the overlap degree, i.e., the closer the two boxes are.
1) 2D IoU: Mapping the 3D detection result box back to a 2D image view.
2) 3D IoU: Directly calculating the IoU of the detection result and the ground truth in 3D space.
3) BEV IoU: Mapping the 3D detection result and the ground truth to a 2D bird's-eye view and then calculating the IoU.

AP is calculated by the following method in this paper. When the steps of the method explained, a 3D bbox is taken as an example for $category = Car$, $difficulty = 1$. Note that $difficulty = 1$ actually refers to $difficulty \leq 1$, including $difficulty = -1$.

*Step 1. Calculate 3D IoU*
Determine if a detected bbox matches a ground truth (gt) bbox (IoU > 0.7).

*Step 2. Select gt bboxes and det bboxes*
For $category = Car$, $difficulty = 1$, gt bboxes is select as bboxes with $category = Car$ and $difficulty \leq 1$; det bboxes is select as predicted bboxes with $category = Car$.

*Step 3. Determine the points $(Pi, Ri)$ on the PR curve corresponding to the score thresholds.*
1) Establish an empty set of scores.
2) For each gt bbox, select the unmatched det bbox with the highest $IoU > 0.7$ as the matched box, mark it as matched, and add its predicted score to the set. If no such det bbox exists, do not add its predicted score to the set.
3) Sort the scores in the set from high to low.
4) Calculate the number of gt bboxes, and compute Recall at 0, 1/40, 2/40, ..., using the scores to form the threshold set.

*Step 4. PR curve and AP calculation.*
For each threshold in the set, calculate a pair $(Pi, Ri)$:
1) Matching: For each gt bbox, select the unmatched det bbox with a prediction $score \geq threshold$ and $IoU > 0.7$ as the matched box, and mark it as matched.
2) Calculate TP: The number of all matched det bboxes. If the corresponding 2D bbox height does not meet the $difficulty \leq 1$ requirement, it is not included in TP.
3) Calculate FN: The number of all unmatched gt bboxes. Even if it is only matched by a det bbox that does not meet the $difficulty \leq 1$ requirement, it is still not counted as FN.
4) Calculate FP: The number of all unmatched det bboxes with a prediction $score \geq threshold$ and height meeting the $difficulty \leq 1$ requirement. For the evaluation of 2D bbox AP, it should be noted that if a det bbox matches a gt bbox labeled as Dontcare, it is not counted as FP (some

Dontcare labeled objects are far away, visible in 2D but not in 3D; therefore, they are labeled in 2D but not in 3D).

5) Calculate Precision $P_i = \frac{TP}{TP+FP}$, and recall: $R_i = \frac{TP}{TP+FN}$

6) PR curve and AP calculation: Based on the pairs $(Pi, Ri)$, plot the curve as shown by the green curve in the figure below, and the area under the curve is recorded as AP. However, in implementation, the evaluation is done using 11 recall positions, as shown by the red line in Fig.2. mAP is the average of all (3) category APs.

AOS is average orientation similarity, to measure the similarity between the detection result and the ground truth orientation. In this article, AOS is also utilized to evaluate the detection accuracy of the model. It is to show the performance of the enhancement.

AOS is calculated by

$$AOS = \frac{1}{11} \sum_{r \in 0,0.1,\dots,1} \max_{\tilde{r}:\tilde{r} \geq r} s(\tilde{r})$$

where, $r$ represents the recall rate of object detection. At recall rate $r$, the orientation similarity $s \in [0,1]$ is defined as the normalization of the cosine distance between all predicted samples and the ground truth. $s$ is calculated by

$$s(r) = \frac{1}{|D(r)|} \sum_{i \in D(r)} \frac{1 + \cos \Delta \theta_i}{2} \delta_i$$

where, $D(r)$ represents the set of all predicted positive samples at recall rate $r$, and $\Delta \theta_i = \theta_i^{gt} - \theta_i^p$ represents the difference between the predicted angle and the ground truth angle for detected object $i$. To penalize multiple detections matching the same ground truth, if detection $i$ has already matched the ground truth, $\delta_i$ is set as 1; otherwise, $\delta_i$ is set as 0.

## 4. Results

Experiments are implemented to explore the impact of varying dropout rates on neural network performance. Fig.3 to Fig.6 depict training and validation losses under different dropout scenarios. The study also evaluates the accuracy performance using AP and AOS metrics, showing differences in performance across the dropout scenarios.

Fig.3 to Fig.6 display the outcomes under various dropout scenarios, exhibiting both similarities and differences. Regarding the similarities, the training loss experiences a significant decline at the initiation of the experiment, followed by a gradual slowdown after approximately 5,000 steps. Furthermore, the trend tends to flatten and escalate after approximately 60,000 steps. The validation loss, calculated using the training dataset without dropout, is nearly identical to the training loss, with the gap between them widening as the dropout value escalates. This is logical since both experiments utilize nearly identical datasets. The difference is that a portion of the convolutional layer is removed during the calculation of the training loss, but not the other. However, the fluctuations seem more significant in the curve representing the validation loss calculated using the training dataset without dropout, particularly after 60,000 steps, where both the training loss and the validation loss calculated without dropout escalate. In terms of the validation loss, it also decreases at the beginning of the experiment, albeit to a lesser extent than the other two types of loss.

Fig.3 displays the results of losses where the dropout rate equals 0. Fig.4 displays the results of losses where the dropout rate equals 0.1. Fig.5 displays the results of losses where the dropout rate equals 0.2. Fig.6 displays the results of losses where the dropout rate equals 0.4.

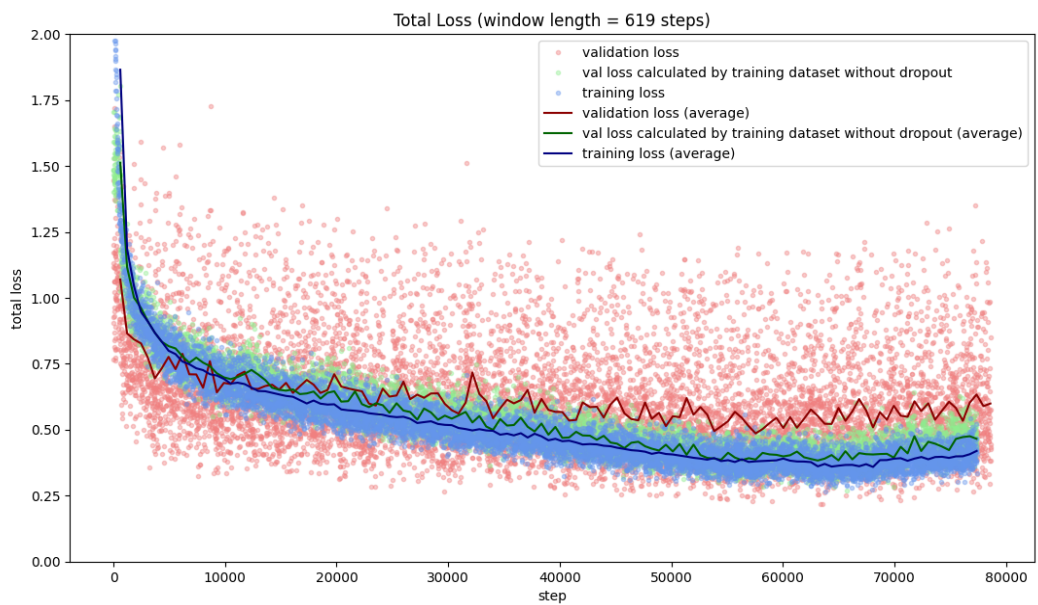

Fig.3　　The results of losses where dropout rate equals 0.

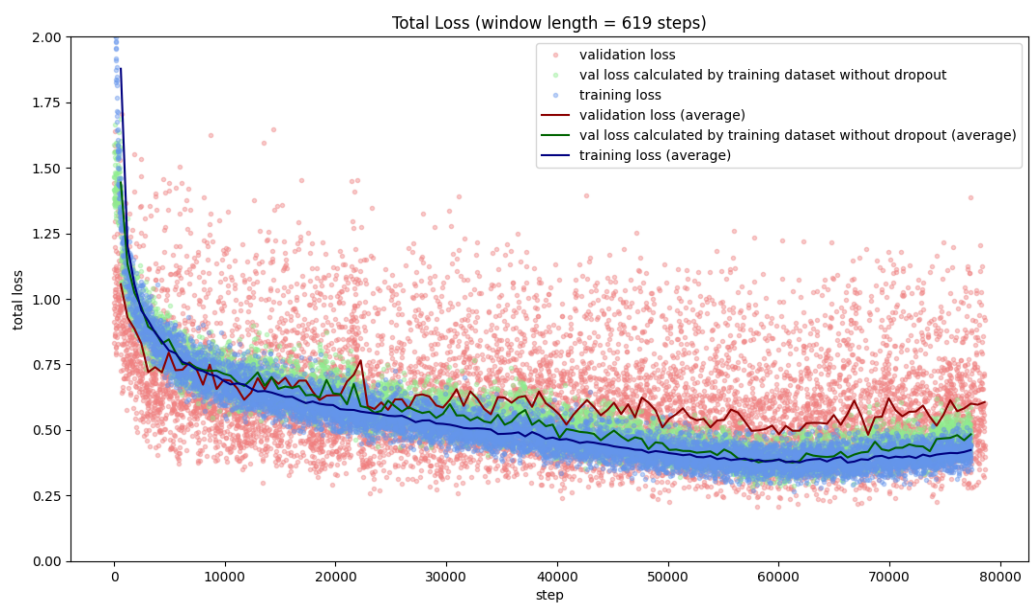

Fig.4　　The results of losses where dropout rate equals 0.1.

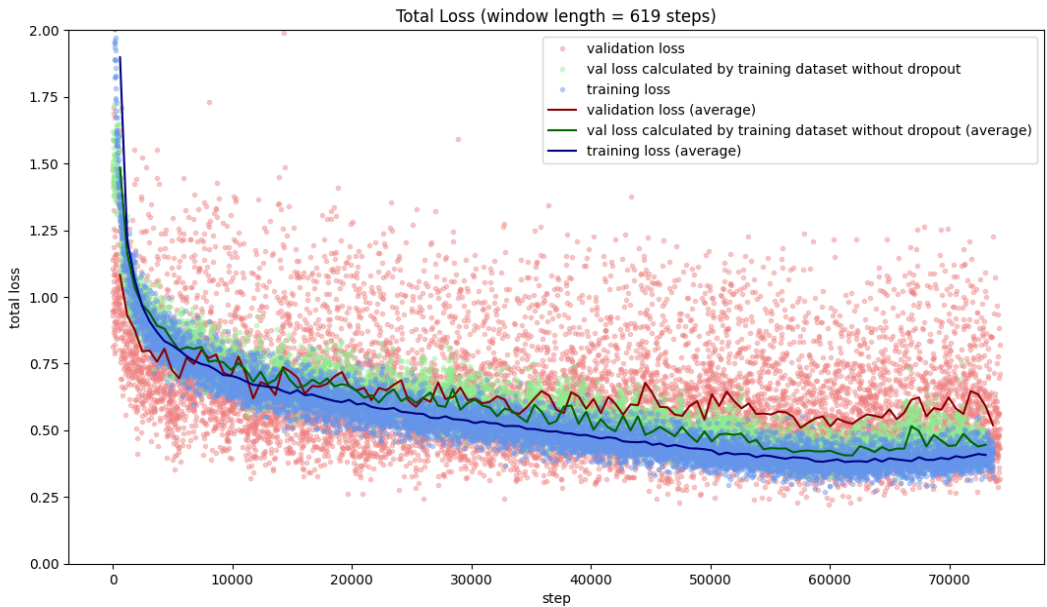

Fig.5　　The results of losses where dropout rate equals 0.2.

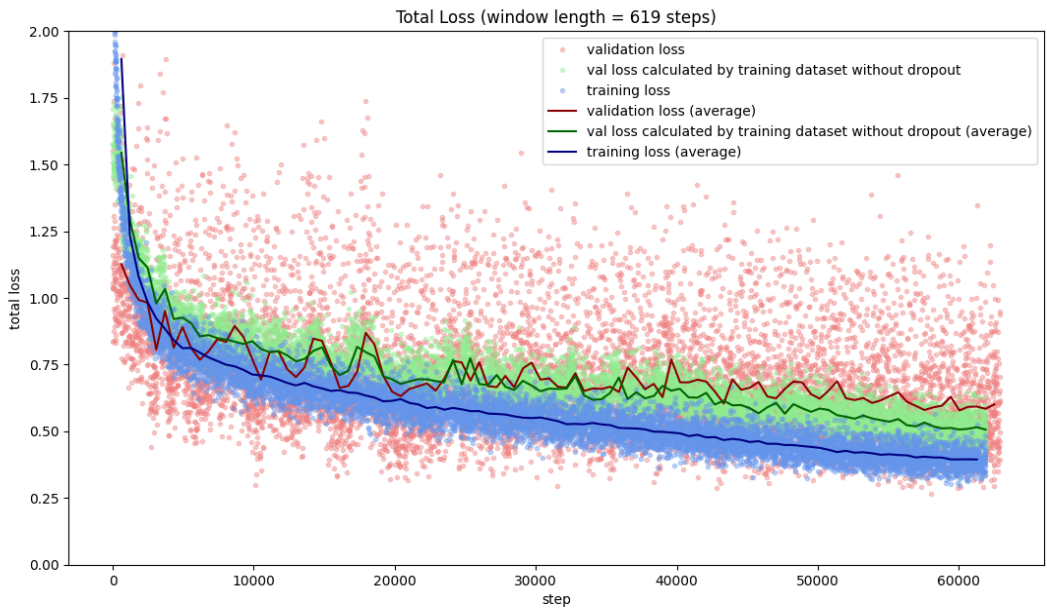

Fig.6　　The results of losses where dropout rate equals 0.4.

The disparity among losses based on varying dropout rates is primarily centered on validation loss. Illustrating the differences, all validation losses are depicted on the same graph, as presented in Fig. 7. As depicted in the figure, the value of validation loss with a dropout rate of 0.4 is notably higher than the other three types. Consequently, we concentrate on analyzing the outcomes with a dropout rate ranging from 0 to 0.2. The three validation losses appear to be closely aligned, with a decline prior to 60,000 steps and an increase thereafter.

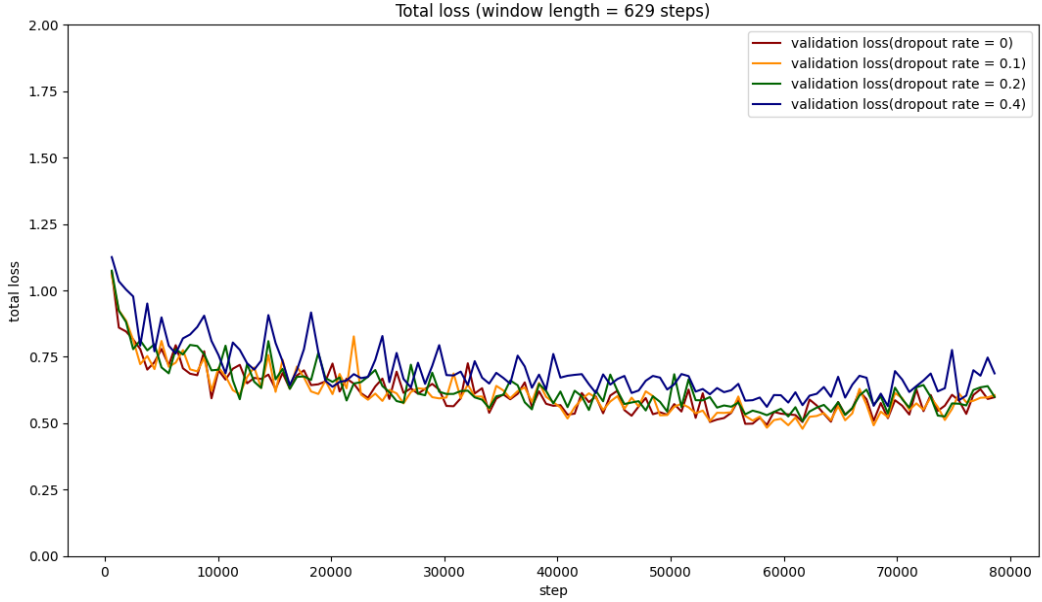

Fig.7　Validation loss based on different dropout rate.

Since the difference among loss curves based on dropout equals 0, 1, and 2, another method of evaluation is utilized to evaluate the accuracy performance of different methods. AP and AOS are calculated by the method introduced in the section of Evaluation Methods, and the results are shown from Table.2 to Table.5.

Table.2　AP BBOX_2D Results based on different dropout rate

| dropout | 0 | | | 0.1 | | | 0.2 | | |
|---|---|---|---|---|---|---|---|---|---|
| difficulty | 0 | 1 | 2 | 0 | 1 | 2 | 0 | 1 | 2 |
| Pedestrian | 61.99 | 57.25 | 55.00 | 58.70 | 55.09 | 52.20 | 55.07 | 52.08 | 49.06 |
| Cyclist | 82.78 | 69.07 | 66.63 | 80.63 | 68.54 | 64.58 | 76.69 | 66.32 | 63.62 |
| Car | 90.64 | 88.11 | 85.69 | 90.22 | 87.67 | 85.04 | 89.76 | 84.84 | 79.68 |
| Overall | 78.47 | 71.48 | 69.11 | 76.52 | 70.43 | 67.28 | 73.84 | 67.74 | 64.12 |

Table.3　AP BBOX_3D Results based on different dropout rate

| dropout | 0 | | | 0.1 | | | 0.2 | | |
|---|---|---|---|---|---|---|---|---|---|
| difficulty | 0 | 1 | 2 | 0 | 1 | 2 | 0 | 1 | 2 |
| Pedestrian | 50.75 | 45.97 | 42.48 | 41.75 | 39.24 | 35.57 | 39.91 | 37.07 | 33.85 |
| Cyclist | 74.15 | 56.70 | 54.19 | 74.35 | 57.08 | 54.33 | 65.86 | 52.68 | 50.38 |
| Car | 77.46 | 72.23 | 66.92 | 74.46 | 65.88 | 64.86 | 62.99 | 56.63 | 53.36 |
| Overall | 67.45 | 58.30 | 54.53 | 63.52 | 54.07 | 51.59 | 56.25 | 48.79 | 45.86 |

Table.4 AP BBOX_BEV Results based on different dropout rate

| dropout | 0 | | | 0.1 | | | 0.2 | | |
|---|---|---|---|---|---|---|---|---|---|
| difficulty | 0 | 1 | 2 | 0 | 1 | 2 | 0 | 1 | 2 |
| Pedestrian | 59.44 | 54.11 | 48.80 | 54.43 | 49.93 | 45.64 | 51.04 | 46.60 | 43.14 |
| Cyclist | 75.34 | 60.01 | 56.75 | 78.54 | 61.68 | 58.36 | 69.85 | 57.45 | 53.62 |
| Car | 90.04 | 84.56 | 79.64 | 89.86 | 84.59 | 84.73 | 90.06 | 84.84 | 79.74 |
| Overall | 74.94 | 66.23 | 61.73 | 74.28 | 65.40 | 62.91 | 70.32 | 62.96 | 58.84 |

Table.5 AOS Results based on different dropout rate

| dropout | 0 | | | 0.1 | | | 0.2 | | |
|---|---|---|---|---|---|---|---|---|---|
| difficulty | 0 | 1 | 2 | 0 | 1 | 2 | 0 | 1 | 2 |
| Pedestrian | 40.50 | 37.66 | 36.01 | **42.17** | **40.23** | **38.30** | 38.57 | 37.42 | 35.31 |
| Cyclist | **81.98** | 66.33 | **63.69** | 79.87 | **66.94** | 62.76 | 75.83 | 64.35 | 61.63 |
| Car | **90.43** | **87.36** | **84.53** | 90.05 | 86.99 | 84.01 | 89.59 | 84.25 | 78.70 |
| Overall | **70.97** | 63.78 | 61.41 | 70.70 | **64.72** | **61.69** | 67.99 | 62.01 | 58.55 |

As shown in the tables, it can be found that both AP and AOS are effected by dropout significantly. AP will not be optimized by increasing dropout, while AOS can be optimized when drop is increased to 0.1.

Compared with dropout equals 0, for detection of pedestrian, 0.1 dropout shows improvement across all difficulties, and 0.2 dropout results in a performance drop across all difficulties. For detection of cyclist, 0.1 dropout results in a slight increase at difficulty 1 but decreases at difficulties 0 and 2, while 0.2 dropout leads to a performance decrease across all difficulties. For detection of car, both 0.1 dropout and 0.2 dropout show a slight decrease in performance across all difficulties, and the performance drop is more significant with 0.2 dropout. For overall performance, 0.1 dropout has mixed effects, with a slight decrease at difficulty 0, a slight increase at difficulty 1 and difficulty 2, while 0.2 dropout consistently results in lower performance across all difficulties.

Consequently, a moderate dropout rate (0.1) might provide some benefits for pedestrian detection and part of cyclist detection but tends to decrease performance for car detection. However, a higher dropout rate (0.2) generally reduces performance across all categories and difficulties.

## 5. Conclusion

The evaluation of the PointPillars model under various dropout rates reveals that dropout significantly affects both the training process and the model's accuracy. While a moderate dropout rate of 0.1 shows some benefits for pedestrian detection and part of cyclist detection, it generally decreases performance for car detection. Higher dropout rates, such as 0.2, consistently reduce performance across all categories and difficulties. Therefore, while dropout can enhance generalization by preventing overfitting, the choice of dropout rate must be carefully balanced to optimize performance in 3D object detection tasks using LiDAR data.

The code we used to train and evaluate the model is available at https://github.com/sun-xiaoxiang/PointPillars-Dropout .

## Acknowledgments

Special thanks to Tom St. John for suggesting this software (https://mlcommons.org/working-groups/benchmarks/automotive/).

preprint arXiv:2003.11755 (2020).